\documentclass[lettersize,journal]{IEEEtran}
\usepackage{amsmath,amsfonts}
\usepackage[table]{xcolor}
\usepackage{array}
\usepackage[caption=false,font=normalsize,labelfont=sf,textfont=sf]{subfig}
\usepackage{textcomp}
\usepackage{stfloats}
\usepackage{url}
\usepackage{mathtools}
\usepackage{verbatim}
\usepackage{graphicx}
\usepackage{pifont}
\def\BibTeX{{\rm B\kern-.05em{\sc i\kern-.025em b}\kern-.08em
    T\kern-.1667em\lower.7ex\hbox{E}\kern-.125emX}}
\usepackage{balance}
\usepackage{cite}
\usepackage{bm}
\usepackage{algorithm}
\usepackage{algpseudocode}
\usepackage{booktabs} 
\usepackage{multirow}
\usepackage{wrapfig}

\def\defn{\,\coloneqq\,}

\def\R{\mathbb{R}}
\def\E{\mathbb{E}}

\def\cbm{{\bm{c}}}
\def\xbm{{\bm{x}}}
\def\zbm{{\bm{z}}}
\def\ybm{{\bm{y}}}

\def\Zbm{{\bm{Z}}}

\def\Sigmabm{{\bm{\Sigma}}}
\def\mubm{{\bm{\mu}}}
\def\thetabm{{\bm{\theta }}}

\def\Ibf{{\mathbf{I}}}

\def\Ncal{{\mathcal{N}}}

\def\Dsf{{\mathsf{D}}}
\def\Rsf{{\mathsf{R}}}

\def\argmin{\mathop{\mathsf{arg\,min}}}
\newcommand{\norm}[1]{\left\lVert#1\right\rVert}

\begin{document}
\title{Multimodal Diffusion Bridge with Attention-Based SAR Fusion for Satellite Image Cloud Removal}

\author{
Yuyang Hu,\thanks{
Yuyang Hu is with the Department of Electrical \& System Engineering, Washington University in St.~Louis, USA. This work was done when Yuyang Hu was an intern at MERL.}
Suhas Lohit, Ulugbek S. Kamilov\thanks{Ulugbek S. Kamilov is with the Departments of Computer Science \& Engineering and Electrical \& System Engineering, Washington University in St.~Louis, USA.}, Tim K. Marks\thanks{Suhas Lohit and Tim K. Marks are with Mitsubishi Electric Research Laboratories (MERL), USA.}
}

\markboth{}%
{How to Use the IEEEtran \LaTeX \ Templates}

\maketitle

\begin{abstract}
Deep learning has achieved some success in addressing the challenge of cloud removal in optical satellite images, by fusing with synthetic aperture radar (SAR) images. Recently, diffusion models have emerged as powerful tools for cloud removal, delivering higher-quality estimation by sampling from cloud-free distributions, compared to earlier methods. However, diffusion models initiate sampling from pure Gaussian noise, which complicates the sampling trajectory and results in suboptimal performance. Also, current methods fall short in effectively fusing SAR and optical data. To address these limitations, we propose Diffusion Bridges for Cloud Removal, DB-CR, which directly bridges between the cloudy and cloud-free image distributions. In addition, we propose a novel multimodal diffusion bridge architecture with a two-branch backbone for multimodal image restoration, incorporating an efficient backbone and dedicated cross-modality fusion blocks to effectively extract and fuse features from synthetic aperture radar (SAR) and optical images. By formulating cloud removal as a diffusion-bridge problem and leveraging this tailored architecture, DB-CR achieves high-fidelity results while being computationally efficient. We evaluated DB-CR on the SEN12MS-CR cloud-removal dataset, demonstrating that it achieves state-of-the-art results.
\end{abstract}


\section{Introduction}
\IEEEPARstart{S}{atellite-based} remote sensing imagery is important in a wide range of fields, such as agriculture monitoring~\cite{grasslandmonitoring} and land use mapping~\cite{yin2021integrating}. One of the biggest challenges is cloud coverage~\cite{background}, which can decrease the image contrast and obscure important details. 

To address this issue, there is growing interest in developing restoration algorithms to recover the lost or distorted signals of satellite images. Early methods for cloud removal in satellite imagery are based on the assumption that cloud-obscured areas share similar statistical and geometric characteristics with the cloud-free regions. By treating cloud removal as an image inpainting problem, these techniques aim to reconstruct missing data by utilizing information from neighboring uncorrupted areas~\cite{chan2001nontexture, maalouf2009bandelet}.

 \begin{figure}
	\centering
	\includegraphics[width=.48\textwidth]{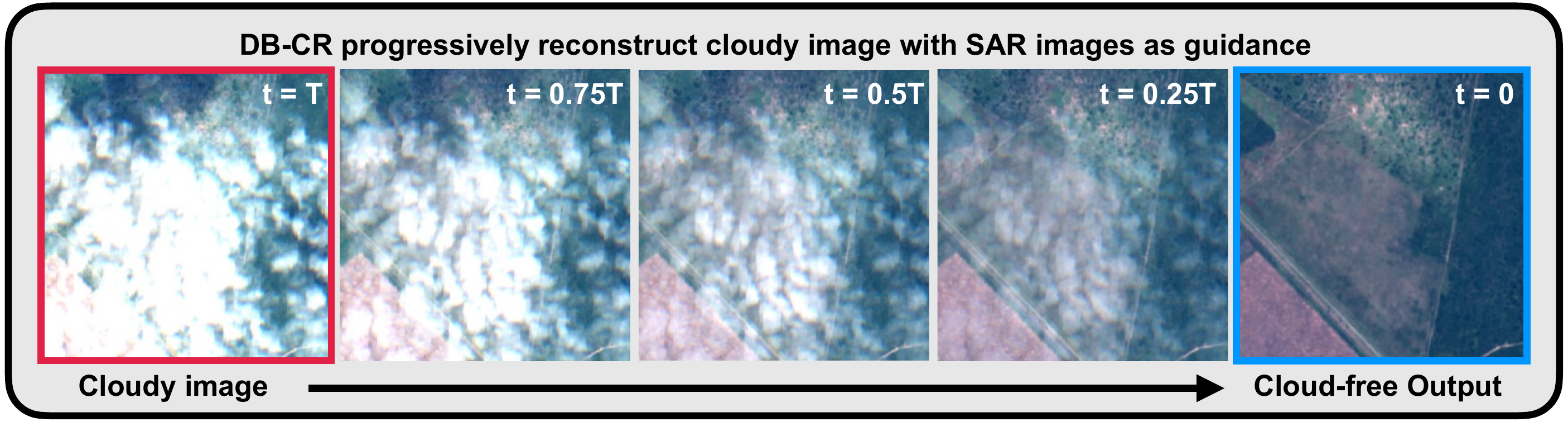}
	\caption{Illustration of DB-CR's iterative process: leveraging SAR data and training a diffusion bridge approach to progressively refine cloud-covered images, reducing cloud cover step by step to produce a clear, cloud-free result.}
	\label{fig:pipeline}
\end{figure}

Deep learning has emerged as a powerful tool for addressing cloud-removal tasks. Instead of explicitly defining an inpainting problem, deep learning methods employ deep neural networks (DNNs) to map cloud-covered images directly to their desired cloud-free counterparts. However, due to the severely ill-posed nature of cloud removal (CR) problems, direct learning of such a mapping can lead to over-smooth predictions. To address this, generative models, particularly conditional Generative Adversarial Networks (cGANs), have shown the potential to produce realistic images~\cite{ctgan, mcgan, spa-gan, stgan, grohnfeldt2018conditional, bermudez2018sar}. The training of cGANs relies on adversarial learning, in which a generator outputs estimated cloud-free images, and a discriminator tries to distinguish these estimates from ground-truth reference images. This adversarial training process helps cGANs to remove cloud coverage and produce images with reasonably realistic texture and details. However, one significant limitation of cGAN-based methods is the instability of their adversarial training process, which can lead to inconsistent performance. 

Some studies explore multi-temporal CR methods, which use a sequences of images captured from the same location at different times. By leveraging the information of other cloud-free or cloudless time points to aid the reconstruction, multi-temporal CR methods achieve improved performance. In many applications, however, multi-temporal data may not be available or easily accessible.

Many cloud-removal techniques rely on synthetic aperture radar (SAR) data to enhance the reconstruction of cloud-obscured pixels, especially when only mono-temporal data (a single image from each location) are available~\cite{ctgan, mcgan, spa-gan, stgan, grohnfeldt2018conditional, bermudez2018sar, dsen2-cr, glf-cr, uncrtaints}. Since SAR is robust to weather conditions and can penetrate clouds~\cite{bamler2000principles}, SAR provides an important supplementary data stream to optical imagery for accurate cloud removal and surface reconstruction. In this work, we focus on multimodal -- SAR + optical image -- data, which allows us to explore cloud removal in scenarios in which temporal information is limited.

Recently, diffusion models (DMs) have emerged as a promising alternative to cGAN-based methods for cloud removal~\cite{ddpm-cr,zhao2023cloud, zou2024diffcr, sui2024diffusion}. DMs contain two Markov processes: a pre-defined forward process that progressively adds noise to clean data, and a learning-based reverse process that attempts to recover the original data from the noisy version. By iteratively adding and removing noise, DMs can learn and sample from the target distribution in a more stable manner. CR methods that use diffusion models represent the current state of the art in cloud removal~\cite{zou2024diffcr, sui2024diffusion}.

Building on the strengths of DMs, a new type of model known as a Diffusion Bridge (DB)~\cite{indi2023, liu2023i2sb, chung2024direct, luo2023refusion, kim2023unpaired} offers a more suitable approach for image restoration tasks when both start (degraded) and end (clean) conditions are precisely defined. Unlike standard DMs, which employ a forward process that progressively adds noise and a reverse process to remove it, DBs explicitly model the transition between two fixed states over time, see Figure~\ref{fig:pipeline}. They thus address a key limitation of standard diffusion models, which often lack sufficient control over the transition and can yield oversmoothed or inconsistent results. By constraining the model to follow these specific states, DBs enhance reconstruction stability and fidelity. 

We hypothesized that these characteristics make DBs particularly well suited to the task of cloud removal, in which the initial (cloudy) and target (cloud-free) states are clearly established. The above considerations led us to develop our proposed method, multimodal Diffusion Bridge for Cloud Removal (DB-CR). Our work makes four key contributions:
\begin{enumerate}
    \item We introduce the \textit{first} application of diffusion bridges to the cloud removal task, DB-CR. Our method, which is specifically designed for multimodal cloud removal, is more stable and accurate than existing approaches in literature, outperforming even the latest methods that are based on  diffusion models.
    \item We propose a novel multimodal diffusion bridge architecture, incorporating a two-branch time-embedded feature extraction block and a cross-modal attention-based fusion block to effectively leverage information from both modalities -- optical and SAR images -- for effective cloud removal.
    \item We perform extensive experiments, which clearly show that DB-CR achieves state-of-the-art performance compared with several existing cloud-removal methods on the widely used SEN12MS-CR benchmark. 
    \item Our method offers the flexibility to adjust the number of inference steps, enabling a customizable trade-off between minimizing a distortion metric like mean-squared error that leads to blurrier outputs and maximizing perceptual performance by preserving sharper details like roads and rivers in the recovered cloud-free optical image\footnote{This trade-off is present in all image restoration algorithms~\cite{blau2018perception}, but using diffusion bridges allows customizing the trade-off point easily} .
\end{enumerate}

 \begin{figure*}
	\centering
	\includegraphics[width=\textwidth]{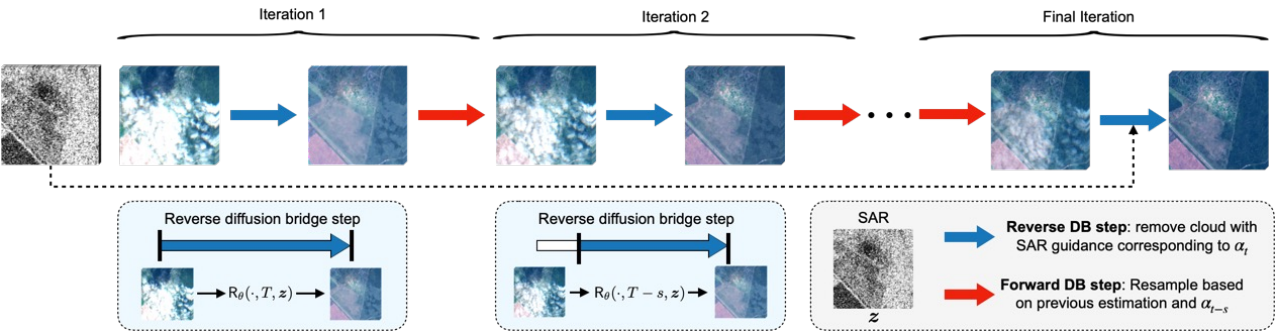}
	\caption{A schematic diagram of our method, illustrating the inference process. Each iteration alternates between a reverse diffusion-bridge step, which estimates the cloud-free state, and a forward diffusion-bridge step, which progresses the state to the next diffusion timestep. The process starts from the cloudy image, progressively refining it toward a cloud-free output. $T$ is the number of diffusion timesteps used in training the diffusion bridge. During inference, $N = \frac{T}{s}$ diffusion timesteps are used, where $s$ is the step size. $\zbm$ refers to the SAR image. }
	\label{fig:pipeline_sampling}
\end{figure*}

\section{Background}
\subsection{Conditional diffusion models for image restoration}
Conditional Diffusion Models (cDM) are designed to learn a parameterized Markov chain to convert a Gaussian distribution into a conditional data distribution $p(\xbm|\cbm)$~\cite{ho2020denoising}. Recently, many cDM-based methods have shown great performance in various image restoration tasks~\cite{wang2022zero, zhu2023denoising}. In the multimodal cloud-removal task, $\xbm$ refers to the target cloud-free optical image, and the condition $\cbm$ refers to the combination of the corresponding cloudy image and SAR image. The forward diffusion process starts from a clean sample $\xbm_0\defn \xbm \sim p(\xbm)$, then gradually adds Gaussian noise to $\xbm_0$ according to the following transition probability:
\begin{equation}
    q(\xbm_t|\xbm_{t-1}) := \Ncal{(\xbm_t; \sqrt{1-\beta_t}\xbm_{t-1},\beta_t\Ibf}),
\end{equation}
where $t\in[0, T]$, $\Ncal(\xbm; \mubm, \Sigmabm)$ denotes the Gaussian probability density function (pdf) over $\xbm$ with mean $\mubm$ and covariance matrix $\Sigmabm$, and $\beta_{t}$ is the noise schedule of the process. By the parameter changes $\alpha_t := 1-\beta_{t}$ and $\bar{\alpha}_t := \prod^{t}_{s=1}\alpha_s$, we can rewrite $\xbm_t$ as a linear combination of $\bm\epsilon_t$ and $\xbm_0$:
\begin{equation}
    \xbm_t = \sqrt{\bar{\alpha}_t}\xbm_0 + \sqrt{1-\bar{\alpha}_t}\bm\epsilon_t,
\end{equation}
where $\bm{\epsilon_t} \sim \Ncal(0, \Ibf)$. This yields a closed-form expression for the marginal distribution of $\xbm_t$ given $\xbm_0$, which simplifies sampling across multiple steps of the Markov chain:
\begin{equation}
    q(\xbm_t|\xbm_{0}) := \Ncal(\xbm_t; \sqrt{\bar{\alpha}_t}\xbm_0, (1-\bar{\alpha}_t)\Ibf).
\end{equation}
The goal of the reverse process is to generate a clean image $\xbm_0$ given a pure Gaussian noise image $\xbm_T$ and the condition $\cbm$. This is achieved by training a neural network $\bm{\epsilon}_\thetabm$, parameterized by $\thetabm$, to reverse the Markov Chain from $\xbm_T$ to $\xbm_0$ one step at a time. The learning is formulated as the estimation of a parameterized Gaussian process
\begin{equation}
    p(\xbm_{t-1}|\xbm_{t}) := \Ncal(\xbm_{t-1}; \bm{\mu}_\thetabm(\xbm_t, t, \cbm), \sigma^2_t\Ibf),
\end{equation}
where the mean $\bm{\mu}_\thetabm(\xbm_t, t, \cbm)$ is only variable we need to estimate, and  $\sigma^2_t = \frac{1-\bar{a}_t}{1-\bar{a}_{t-1}}\beta_t$. In cDM, $ \bm{\epsilon}_\thetabm $ is trained to predict the noise $ \bm{\epsilon_t} $ added during the forward process by minimizing the following expected loss:

\begin{equation}
\label{Eq:dm_training}
\mathcal{L}(\thetabm) = \E_{\bm{\epsilon_t}, \xbm_t, t} \left[ \norm{\bm{\epsilon}_\thetabm(\xbm_t, t, \cbm) - \bm{\epsilon_t}}_2^2 \right].
\end{equation}

The estimated noise is then used to reparameterize the sampling from the conditional distribution:

\begin{equation}
    \xbm_{t-1} = \frac{1}{\sqrt{\alpha_t}} \left( \xbm_t - \frac{\beta_t}{\sqrt{1 - \bar{\alpha}_t}} \bm{\epsilon}_\thetabm(\xbm_t, t, \cbm) \right) + \sigma_t \bm{z},
\end{equation}
where $ \bm{z} \sim \Ncal(\mathbf{0}, \Ibf) $. By iteratively applying this reverse process from $t = T$ to $t = 0$, the trained cDM can reconstruct the clean sample $\xbm_0$ by progressively sampling from the learned conditional distribution.

\subsection{Conditional diffusion models for cloud removal}
Recent works have employed cDMs to address cloud removal tasks, leveraging the power of diffusion models to produce high-quality cloud-free images, such as in DDPM-CR~\cite{jing2023denoising} and DiffCR~\cite{zou2024diffcr}. In DDPM-CR, the authors directly applied cDMs~\cite{ho2020denoising} to the cloud removal problem by concatenating the cloudy image as a condition with the input of the diffusion model. DiffCR further advanced this by introducing a more efficient network architecture, incorporating an additional encoder to better utilize conditional inputs, and achieving faster inference through an optimized ODE solver rather than the original DDPM solver. These cDM-based methods have demonstrated substantial improvements in reconstructing cloud-free satellite imagery, particularly in scenarios with varying degrees of cloud coverage.

However, despite promising results, the stochastic nature of the reverse process can sometimes lead to inconsistencies and artifacts, especially in areas where the model needs to be more precise in reconstructing details. This motivated us to consider the use of more deterministic approaches. In particular, we focused on the diffusion bridge framework. In our approach, the diffusion bridge explicitly controls the transformation between the cloudy and cloud-free states, providing a more stable and effective solution to the cloud removal problem.

\subsection{Diffusion bridges for image restoration}

Generation in DMs is achieved by transforming noise into a target distribution through a series of denoising steps. The diffusion bridge (DB) method generalizes this approach by modeling a stochastic process that connects two fixed states, where the initial state is not limited to a Gaussian distribution. In the context of image restoration, DBs often involve constructing a probabilistic trajectory between two states (e.g., degraded and clean images) using an optimal transport formulation~\cite{indi2023, liu2023i2sb, chung2024direct}. This approach leverages the inherent similarity between these states to efficiently learn mappings, leading to improved performance and faster inference~\cite{indi2023, liu2023i2sb, chung2024direct, luo2023refusion}. For example, InDI~\cite{indi2023} proposed a diffusion bridge framework that achieved state-of-the-art results across various real-world image restoration tasks, including image super-resolution, image deblurring, and JPEG artifact removal, while achieving significantly faster performance compared to traditional DMs. Moreover, recent work~\cite{hu2024stochastic} demonstrates that a DB-based restoration model offers superior robustness to distributional shifts between training and testing datasets compared with DM-based method. By learning to restore a diverse range of degraded states, DB models provide enhanced stability to out-of-distribution data. While DBs have demonstrated strong performance in single modality settings, it has neither been extended to multimodal restoration tasks nor applied to the specific challenges of cloud removal.

A diffusion bridge explicitly models the transition from a degraded image state to a clean image state through an Optimal Transport (OT) framework. Give a tuple $(\xbm_0, \ybm) \sim p(\xbm, \ybm)$, where $ \ybm $ represents the degraded image and $\xbm_0$ represents the target clean image, the forward process in DB begins from $ \xbm_0 $ and transforms it towards $ \ybm $, with each intermediate state $ \xbm_t $ representing a linear mixture of both the corrupted and target images. This process is defined as:
\begin{equation}
    \xbm_t = (1 - \alpha_t) \xbm_0 + \alpha_t \ybm, \quad \alpha_t \in [0, 1],
\end{equation}
where $ \alpha_t $ is a monotonically increasing function as t goes from $0$ to $T$. Specifically, $ \alpha_0 = 0 $ and $ \alpha_T = 1 $.  As $ t $ progresses from 0 to $ T $ in the forward process, $ \xbm_t $ transitions smoothly from the clean image $ \xbm_0 $ to the degraded image $ \ybm $, creating a well-defined trajectory that deterministically blends the two image states.  Conversely, in the reverse process, as $ t $ progresses from $ T $ to $ 0 $, $ \xbm_t $ evolves from $ \ybm $ back to $ \xbm_0 $ through a deterministic trajectory, progressively removing the degradation and reconstructing the original clean image. 

The reverse process in the DB framework reconstructs the clean image from the degraded observation through a multi-step sampling strategy, utilizing an Optimal Transport Ordinary Differential Equation (OT-ODE). Unlike traditional end-to-end cloud removal models that attempt to estimate a clean image in a single pass, the Diffusion Bridge breaks down the transformation into a sequence of smaller, more manageable steps, each guided by the OT-ODE. The OT-ODE is described by:
\begin{equation}
    \frac{d \xbm_t}{d t} = v_t(\xbm_t|\xbm_0), \quad \text{where} \quad v_t(\xbm_t|\xbm_0) = \alpha_t (\xbm_0 - \xbm_t),
\end{equation}
where $ v_t(\xbm_t|\xbm_0) $ represents the optimal velocity field that guides $ \xbm_t $ towards the target state $ \xbm_0 $. To approximate this velocity field, a neural network $ \Rsf_\theta(\xbm_t, t) $ is trained to determine the optimal direction for each step of the reverse evolution. The training objective for the network is formulated as:
\begin{equation}
    \theta^\ast = \argmin_\theta \E_{t, \xbm_t, \xbm_0} \left[ \norm{\Dsf_\theta(\xbm_t, t) - v_t(\xbm_t|\xbm_0)}_2^2 \right],
\end{equation}
where $ \Dsf_\theta(\xbm_t, t) $ predicts the optimal velocity that guides $ \xbm_t $ back to the clean target state $ \xbm_0 $. The multi-step reverse sampling provides incremental updates toward the optimal transport solution, minimizing transport costs and preserving structural details. The controlled velocity at each stage ensures a smooth transition, reducing artifacts and oversmoothing, which are commonly observed in ill-posed problems. By decomposing the reconstruction into multiple steps, the diffusion bridge framework addresses the complexity of difficult restoration tasks by solving progressively simpler sub-problems, showing state-of-the-art performance in various image restoration tasks (e.g., image super-resolution, image deblurring, and image inpainting).

Existing diffusion bridge frameworks have primarily been developed for single-modality image restoration tasks. They have not been applied to or validated on challenging multimodal problems such as cloud removal. In this work, we introduce a new multimodal diffusion brige framework, DB-CR, which is specifically tailored for cloud removal. Not only is the first time that a diffusion bridge has been applied to the task of cloud removal, but as far as we know it is also the first multimodal diffusion bridge. DB-CR incorporates a novel attention fusion mechanism and an efficient dual-branch backbone to effectively integrate structural information from SAR data with fine spectral details from optical imagery.

Unlike prior methods, which lack the capacity to handle multimodal inputs, DB-CR leverages the complementary strengths of SAR for structural information and optical imagery for fine details, achieving state-of-the-art performance. By demonstrating DB-CR on cloud removal---a highly ill-posed multimodal problem---this work sets a new benchmark and establishes the utility of diffusion bridges for previously unexplored multimodal domains.

In the following sections, we detail the innovations in DB-CR and demonstrate that its multimodal diffusion bridge formulation advances the state of the art cloud-removal performance.

 \begin{figure*}
	\centering
	\includegraphics[width=\textwidth]{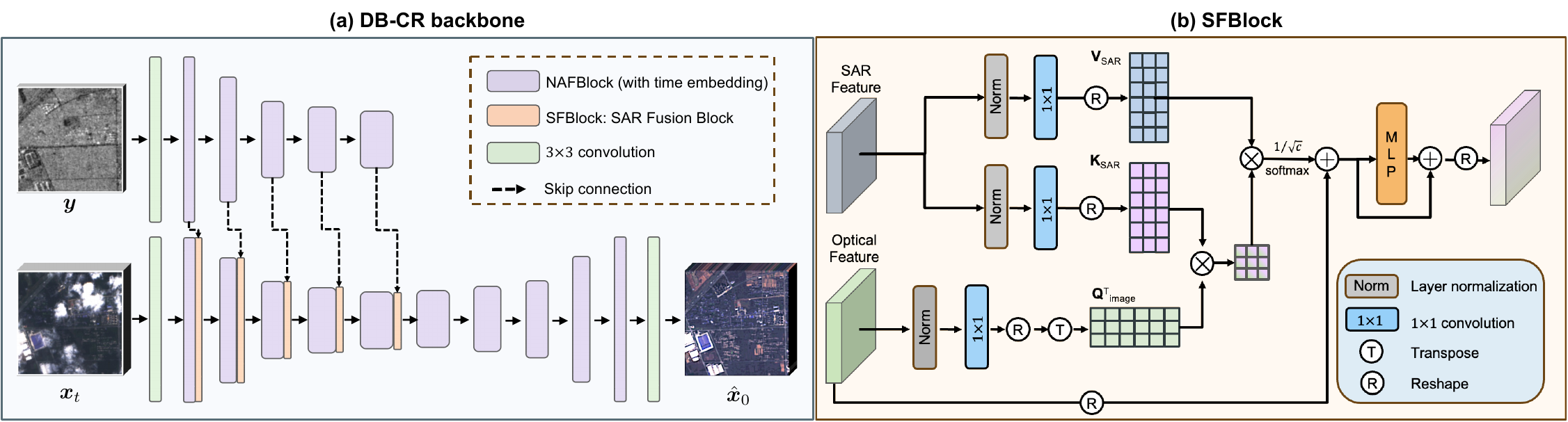}
	\caption{The architecture of the DB-CR backbone network. (a) This two-branch design combines a U-Net restoration branch (bottom), which utilizes NAFBlocks (purple rectangles) for efficient restoration, and a SAR feature extraction branch (top). The network utilizes SFBlocks (orange rectangles) for feature fusion across branches. (b) Detailed architecture of each SFBlock.}
	\label{fig:method}
\end{figure*}

\section{Proposed Method}
\label{Sec:Method}
\subsection{Problem definition}
This work focuses on the SAR-conditioned cloud removal problem. The objective is to generate a cloud-free optical satellite image, $\xbm \in \R^{C, H, W}$, from a single observed cloudy optical image, $\ybm \in \R^{C, H, W}$, and the corresponding SAR image $\zbm \in \R^{C', H, W}$. In our experiments, $C = 13$ is the number of multispectral channels in each optical image, $C'=2$ is the number of channels in the SAR image, and $H$ and $W$ are the images' height and width, respectively.

\subsection{Multimodal Diffusion ridge framework for Cloud Removal (DB-CR)} 
Given a pair of spatially aligned optical images, one cloudy and one cloud-free, we define the forward process of the diffusion bridge (DB) at each discrete time step as 
\begin{equation}
    \xbm_t = (1 - \alpha_t)\xbm_0 + \alpha_t \ybm, \quad t \in \{0,1,\ldots, T\},
\end{equation}
where $\xbm_t$ represents an intermediate image transitioning from the cloud-free image $\xbm$ (at $t=0$) to the cloudy image $\ybm$ (at $t=T$). The weighting factor $\alpha_t$ is a function that increases monotonically from 0 to 1 as $t$ goes from $0$ to $T$, ensuring a smooth and deterministic trajectory between the two image states.

The reverse process aims to recover the clean image $\xbm$ from the cloudy image $\ybm$ by inverting this DB forward process. To accomplish this, we train a neural network $\mathcal{R}_\theta$ to predict the clean image at each time step $t$, given the corrupted intermediate image $\xbm_t$ and the aligned SAR reference image $\zbm$.
The training objective is defined as:
\begin{equation}
\label{eq:train_ode_loss}
    \theta^\ast  = \argmin_\theta \mathbb{E}_{t \sim \{0, 1, \dots, T\}
}  \left\| \Rsf_\theta(\xbm_t, t, \zbm) - \xbm_0 \right\|_1,
\end{equation}
where the model directly learns to predict $ \xbm_0$ at each time step. Here, we use the mean absolute error (MAE) loss function, which has demonstrated good performance for the cloud removal task~\cite{dsen2-cr}.
In this formulation, $ \Rsf_\theta $ learns to predict the minimum mean absolute error (MMAE) estimation, $\E[\xbm_{0}|\xbm_{t}]$, of true cloud-free image $\xbm_0$ from the intermediate corrupted input $\xbm_t$. The pseudocode for this training process is provided in Algorithm~\ref{alg:training}.

Once the network is trained, we define a reverse process to reconstruct the cloud-free image $ \xbm_0 $ from the cloudy observation $ \xbm_T $. As $t$ is decreased from $T$ (initial step) to 0 (final step), $\alpha_t$ progressively decreases from 1 to 0, and the reverse process incrementally refines the image, transforming $ \xbm_T $  back to  $ \xbm_0 $. 

Given a predefined number of function evaluations (NFE), we uniformly select $N$ timesteps from the original diffusion process, by setting an appropriate step size $s$. Following this schedule of diffusion timesteps, the process starts with the original cloudy image $\ybm$ as the initial input and iteratively transforms it toward a cloud-free state. At each iteration, the pre-trained model $\Rsf_\theta$ computes the cloud-free estimate $\xbm_{0|t}'$ based on the current intermediate image $\xbm_t$, time step $t$, and auxiliary SAR data $\zbm$. Using Equation~\eqref{eq:inference_ode}, the next intermediate image $\xbm_{t-1}$ is resampled by combining $\xbm_{0|t}'$ with $\xbm_t$.
\begin{equation}
\label{eq:inference_ode}
    \xbm_{t-s} = \left(1 - \frac{\alpha_{t-s}}{\alpha_t}\right)\xbm_{0|t}' + \left(\frac{\alpha_{t-s}}{\alpha_t}\right) \xbm_t, \quad s < t,
\end{equation}
where $s$ is the step size of the reverse process and $\xbm_{0|t}'$ represents the cloud-free prediction based on the intermediate image  $\xbm_t$, i.e., $\xbm_{0|t}' = \E[\xbm_{0}|\xbm_{t}]$. The overall inference pseudocode is given in Algorithm~\ref{alg:inference}. As illustrated in Figure~\ref{fig:pipeline_sampling}, this repeated resampling progressively reduces cloud cover, ultimately producing the final cloud-free image.

The multi-step generation approach effectively addresses the challenge of over-smoothing, which is common in ill-posed problems like cloud removal. Unlike single-step methods that directly estimate cloud-free images and often produce averaged outputs lacking fine details, this approach decomposes the task into a series of simpler sub-problems under a diffusion bridge formulation. At each step, the model incrementally refines a less degraded image, progressively removing the cloud cover. This iterative refinement not only mitigates over-smoothing but also preserves finer details. Additionally, the stepwise process allows for more effective integration of SAR information at each stage, leading to sharper and more accurate cloud-free reconstructions.

\begin{algorithm}[h]
  \caption{DB-CR (Training)} 
  \label{alg:training}
  \begin{algorithmic}[1]
     \Require  \text{Training dataset:} $p(\xbm,\ybm, \zbm)$, Training diffusion timesteps $T$, $\{{\alpha}_t\}_{t=0}^T$
    \State \textbf{repeat:} 
    \State \quad $\xbm_0,\ybm, \zbm \sim p(\xbm,\ybm, \zbm)$, $t \sim \{0, 1, \dots, T\}$
    \State \quad  $\xbm_t = (1 - \alpha_t)\xbm_0 + \alpha_t \ybm$
    \State \quad Take gradient descent step on \par   \quad $\nabla_\theta \left\| \Rsf_\theta \left(\xbm_t, t, \zbm \right) - \xbm_0 \right\|_1$ 
    \State \textbf{until convergence}
    
\end{algorithmic}
\end{algorithm}

\begin{algorithm}[h]
  \caption{DB-CR (Inference)} 
  \label{alg:inference}
  \begin{algorithmic}[1]
     \Require Cloudy optical and SAR images: $(\ybm, \zbm)$, Trained DB-CR model $\Rsf_\theta$,Training diffusion timesteps $T$, NFE steps $N$  
     \State \text{Get step size:} $s = \frac{T}{N}$

     \State \textbf{Initial:} $\hat{\xbm}_{T} = \ybm $
    \For{$t = T$ to $0$ with step size $s$}

    \State \quad  $\xbm_{0|{t}}' = \Rsf_\theta({\xbm}_{t}, t, \zbm) $

    \State \quad $\xbm_{t-s} \gets \left(1-\frac{{\alpha}_{t-s}}{{\alpha}_{t}}\right)\xbm_{0|{t}}' +  \frac{{\alpha}_{t-s}}{{\alpha}_{t}}\xbm_{t}$
    \EndFor
\State \textbf{return} $\xbm_{0|{t}}'$
    
\end{algorithmic}
\end{algorithm}

\subsection{SAR-enhanced diffusion bridge backbone}

To enhance cloud removal performance and improve feature extraction from SAR data, we propose a novel two-branch diffusion bridge backbone, consisting of a SAR feature extraction branch and an optical image restoration branch, as illustrated in Figure \ref{fig:method}. The backbone is built using two main components: (1) a time-embedded Nonlinear Activation-Free Network (NAFNet) block, and (2) a cross-attention based SAR Fusion Block (SFBlock). The SFBlock enables multi-level fusion of features from both the SAR extraction branch and the optical image restoration branch.

\subsubsection{Two-branch backbone for DB-CR}
As shown in Figure~\ref{fig:method}, the proposed DB-CR backbone has two branches:
a NAFNet-based U-net architecture for image restoration (bottom row) and a NAFNet encoder architecture for SAR feature extraction (top row). This dual-branch design enables dedicated feature extraction for each modality, allowing the network to more efficiently learn modality-specific features.

Traditional diffusion bridge architectures often employ the U-Net using residual blocks and attention mechanisms such as channel-attention and self-attention~\cite{liu2023i2sb, chung2024direct}. While effective in capturing fine details, these components significantly increase computational cost, which can slow down the inference process, particularly in iterative, multi-step generation required by diffusion bridge methods.

To address this drawback, we adopt a more efficient approach by incorporating Nonlinear Activation-Free Blocks (NAFBlocks)~\cite{nafnet, luo2023refusion} into our architecture, thus introducing a more efficient alternative to the commonly used ResBlock backbone. Typically, diffusion bridge frameworks for image restoration, as seen in\cite{liu2023i2sb, chung2024direct}, employ a ResBlock backbone that combines residual and attention blocks to enhance feature representation. However, this setup can be computationally expensive due to the added complexity of attention mechanisms and activation layers. In contrast, NAFBlocks take a streamlined approach by excluding attention mechanisms and activation functions, focusing solely on linear transformations, which significantly reduces the computational overhead.  Each NAFBlock consists of two parts: (1) a mobile convolution module (MBConv) based on point-wise and depth-wise convolution with channel attention, and (2) a feed-forward network (FFN) module that has two fully connected layers. The LayerNorm (LN) layer is added before both MBConv and FFN, and a residual connection is employed for both modules. We denote the intermediate feature maps as $\Zbm_{\text{MBConv}}$ and $\Zbm_{\text{FFN}}$. For a input feature map $\textbf{X}$, the whole process is formulated as:
\begin{equation}
\begin{aligned}
        &\Zbm_{\text{MBConv}} = \text{MBConv(LN(\textbf{X}))} + \textbf{X} \\
    &\Zbm_{\text{FFN}} = \text{FFN}(\text{LN}(\Zbm_{\text{MBConv}})) + \Zbm_{\text{MBConv}}\\
\end{aligned}
\end{equation}
Another special part of NAFBlocks is that they use SimpleGate units to replace nonlinear activation functions (such as ReLU or GELU) after the FFN module. Given an input feature map generated by FFN module $\textbf{X} \in \R^{H\times W\times C}$, SimpleGate first splits the input into two features $\textbf{X}_1, \textbf{X}_2 \in \R^{H\times W\times C/2}$ along the channel dimension.  It then computes the output using element-wise multiplicative gating:
\begin{equation}
    \text{SimpleGate}(\textbf{X}) = \textbf{X}_1 \odot \textbf{X}_2
\end{equation}

While NAFBlocks were originally designed for general RGB image restoration, their ability to streamline feature extraction without sacrificing performance makes them particularly suitable for our multi-spectral cloud removal task. By introducing the NAFBlocks, we reduce the reliance on heavy attention mechanisms and activation functions, achieving a balance between efficiency and accuracy in cloud-free image prediction. This high-efficiency block addresses the fundamental challenge of high computational demand in multi-step restoration, which is crucial in diffusion bridges, resulting in faster inference while preserving image quality.

\subsubsection{SAR-image cross-modal attention fusion}
To effectively fuse the multimodal information from both the SAR and optical image branches, we employ a cross-modal attention mechanism in the SAR Fusion Block (SFBlock), inspired by the network structures in~\cite{chen2024motion, sun2022event}. This block enables the model to capture the complementary nature of SAR and optical data by aligning and integrating features at multiple levels. SAR data, which is robust to atmospheric conditions, provides structural and texture information, while the optical data contributes color and fine details. By employing multi-head cross-attention between these modalities, the SFBlock ensures that the final reconstruction benefits from the strengths of both data sources, ultimately leading to improved cloud removal and image restoration performance.

The SFBlock operates by applying a cross-modal attention mechanism similar to self-attention but with a key difference: the queries ($\mathbf{Q}_{\text{O}}$) are derived from the optical image branch, while the keys ($\mathbf{K}_{\text{S}}$) and values ($\mathbf{V}_{\text{S}}$) come from the SAR branch. In this setup, $h$ and $w$ represent the height and width of the feature maps, and $c$ denotes the number of channels. Initially, both input features have dimensions $ h \times w \times c $, but after passing through normalization and $1 \times 1$ convolution layers, they are reshaped to $ hw \times c $, where $ hw $ represents the flattened spatial dimensions. The cross-modal attention mechanism selectively emphasizes relevant SAR features using guidance from the optical image content, leveraging the unique characteristics of each modality. For each attention head, this process is expressed as:
\begin{equation}
\text{Attention}(\mathbf{Q}_{\text{O}}, \mathbf{K}_{\text{S}}, \mathbf{V}_{\text{S}}) = \mathbf{V}_{\text{S}} \cdot \text{softmax}\left(\frac{\mathbf{Q}_{\text{O}}^T \mathbf{K}_{\text{S}}}{\sqrt{c}}\right),
\end{equation}
where $\mathbf{Q}_{\text{O}}^{\mathrm T} \mathbf{K}_{\text{S}} \in \mathbb{R}^{c \times c}$ represents the similarity scores between optical and SAR features. These scores are computed by taking the transpose of the queries, $\mathbf{Q}_{\text{O}}^{\mathrm T}$ (dimension $c \times hw$), and multiplying it by $\mathbf{K}_{\text{S}}$ (dimension $hw \times c$), yielding a matrix in $\mathbb{R}^{c \times c}$. The resulting attention map in $\mathbb{R}^{c \times c}$ is then applied to $\mathbf{V}_{\text{S}}$ (dimension $hw \times c$) to produce the final attended features in $\mathbb{R}^{hw \times c}$. This channel-wise attention significantly reduces spatial complexity from $O(h^2 w^2)$ to $O(c^2)$, making the operation more efficient without sacrificing performance. Finally, the output of the attention operation is added to the input
image features to produce  $ \Zbm_{\text{sum}}$, which is subsequently passed through a residual-connected multi-layer perceptron (MLP):
\begin{equation}
    \Zbm_{\text{output}} = \Zbm_{\text{sum}} + \text{MLP}(\Zbm_{\text{sum}}).
\end{equation}
After reshaping $\Zbm_{\text{output}}$ back to $ c \times h \times w $,  it serves as the final output of each attention head within the SFBlock. To produce the final output of the entire SFBlock, the outputs from all attention heads are concatenated along the channel dimension. A final $1\times1$ convolution layer is then applied as a linear projection to integrate information across the heads and ensure that the output tensor retains the same channel dimension as the input.

By integrating the efficiency of NAFBlocks with the advanced multimodal fusion capabilities of the SFBlock, our two-branch diffusion bridge backbone is designed for effective cloud removal. This dual-branch architecture combines the structural strengths of the SAR feature extraction branch with the fine details from the optical image restoration branch. Our approach addresses the computational challenges of multi-step restoration while ensuring accurate cloud-free reconstructions that preserve both spatial and spectral integrity.

\section{Experiments}
\subsection{Dataset}
We conduct our experiments using the SEN12MS-CR dataset~\cite{schmitt2019sen12ms}, a multimodal, mono-temporal dataset specifically curated for cloud removal tasks in remote sensing. SEN12MS-CR provides paired and co-registered data from SAR and multi-spectral optical imagery, collected from the Sentinel-1 and Sentinel-2 satellites of the European Space Agency’s Copernicus mission respectively. The optical data contain 13 spectral bands, including visible (RGB), near-infrared (NIR), and shortwave infrared (SWIR) bands. The SAR data contain 2 channels containing VV and VH polarized images. The dataset contains observations covering 175 globally distributed regions of interest recorded in one of four seasons throughout the year of 2018. The full images are divided into a total of 122,218 patch triplets, each patch of size $256\times256$. The training, validation, and test sets contain 114056,
7176, and 7899 examples, respectively, following the dataset splits provided in~\cite{uncrtaints}.
 
Before the images are input into the network, the images undergo preprocessing steps to ensure quality and numerical stability, which are the same as in~\cite{meraner2020cloud}. The Sentinel-2 optical bands are clipped to the range [0, 10,000], then scaled to the range [0, 1]. Similarly, the SAR data are clipped, with VV and VH polarizations being clipped to the range [-25, 0] and [-32.5, 0], respectively. To further align the SAR and optical data distributions, the Sentinel-1 SAR values are shifted into the positive domain and scaled to the range [0, 1], matching the scaled optical data values.

\subsection{Implementation}

\begin{figure}
  \centering
    \includegraphics[width=0.25\textwidth]{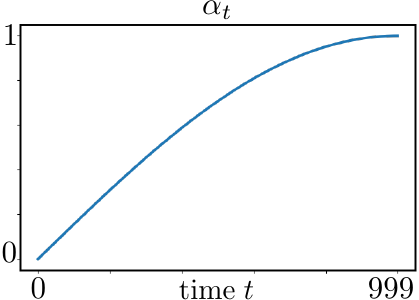}
    \caption{$\alpha_t$ scheduling for DB-CR, which follows sine-based distribution.
}
  
  \label{fig:alpha_t}
\end{figure} 

All experiments were implemented using PyTorch 1.13.1 with CUDA 12.6. For training, we set the number of timesteps for the diffusion bridge process to 1000, and the scheduling of $\alpha_t$ corresponding to each timestep is shown in Figure~\ref{fig:alpha_t}. The training process was conducted over 50 epochs with a batch size of 4, using the Adam optimizer with a learning rate of $5 \times 10^{-5}$. We evaluate both the distortion and perceptual performance of our proposed method, as well as several baseline methods. In the backbone architecture, the SAR and optical branches adopt channel dimensions of [22, 44, 88, 176] at each level. The encoder for both branches is configured with [1, 1, 1, 28] NAFBlocks across levels, while the decoder in the optical branch is designed with [1, 1, 1, 1] NAFBlocks. In both branches, before the NAFBlocks a single $1 \times 1$ convolutional layer is used which maps from 13 channels to 22 channels for the optical image branch, and from 2 channels to 22 channels for the SAR image branch. The outputs from each level of the encoder in both branches serve as inputs to the SFBlock. The SFBlock utilizes an increasing number of attention heads across levels, configured as [1, 1, 2, 4]. The MLP inside consists of two fully connected layers with GELU activation functions, where the hidden dimension is configured to be twice the size of the input dimension.

\subsection{Evaluation metrics}
For evaluating the restoration quality of the reconstructed cloud-free images, we employ a set of widely used quantitative metrics: Peak Signal-to-Noise Ratio (PSNR, in dB), Structural Similarity Index Measure (SSIM)~\cite{ssim}, Mean Absolute Error (MAE), and Spectral Angle Mapper (SAM)~\cite{kruse1993spectral}. PSNR measures the pixel-wise similarity by comparing the reconstructed image to the ground truth, with higher values indicating less distortion. SSIM captures structural similarity by evaluating structural information and luminance consistency. MAE calculates the average absolute difference between predicted and true pixel values, highlighting overall error, while SAM, measured in degrees, quantifies angular deviation in spectral information, important for preserving spectral fidelity. These metrics are calculated using the code provided by UnCRtainTS\footnote{\url{https://github.com/PatrickTUM/UnCRtainTS}}~\cite{uncrtaints}.

To measure perceptual performance, we use Learned Perceptual Image Patch Similarity (LPIPS)~\cite{lpips} and Fréchet Inception Distance (FID)~\cite{fid}. LPIPS is a perceptual metric that assesses similarity based on deep feature activations from neural networks, offering insight into the perceptual closeness of images. FID compares the distribution of deep features between sets of generated and real images, with lower values indicating closer alignment to the original image distribution. Both LPIPS and FID are essential for evaluating perceptual realism, as they reflect how closely reconstructed images resemble ground truth when viewed by a human observer. As the backbone neural networks for both LPIPS and FID are trained on RGB images, we compute these metrics exclusively on the RGB channels of the cloud-free estimations that are generated by the baseline methods and our proposed approach.

To measure model efficiency, we report the parameter count (Params), which represents the model’s size and potential memory footprint, and the number of multiply–accumulate operations (MACs), a measure of computational complexity. These values provide insight into model scalability and resource requirements, particularly useful for deployment in resource-constrained environments. Calculations are performed using the open-source THOP repository\footnote{\url{https://github.com/ultralytics/thop}}.

\begin{table*}[hbt!]
\centering
\caption{Comparison of DB-CR (using NFE=1) with state-of-the-art cloud-removal methods on SEN12MS-CR Dataset. The results clearly show that DB-CR achieves best performance with respect to all the reconstruction evaluation metrics while being computationally efficient.}
\renewcommand\arraystretch{1.25}
\setlength{\tabcolsep}{3mm}{
\begin{tabular}{ccccccccc}
\toprule
Methods & PSNR(dB) $\uparrow$ & SSIM $\uparrow$ & MAE $\downarrow$ & SAM (deg) $\downarrow$ & FID $\downarrow$ & LPIPS $\downarrow$ & Parameters ($\times 10^6$) $\downarrow$ & MACs ($\times 10^9$) $\downarrow$ \\ \hline
        DSen2-CR~\cite{dsen2-cr} &30.29         &0.878      &0.034	     &7.008     &80.37               &0.431    &18.95     &1241.8      \\
        GLF-CR~\cite{glf-cr} &31.85          &0.896      &0.020     &6.722     &87.63               &0.452   &14.83     &245.28       \\
        UnCRtainTS~\cite{uncrtaints} &31.28          &0.887      &0.023     &6.899     &85.94               &0.410     & \textbf{2.02}     &107.57    \\
        DiffCR~\cite{zou2024diffcr} &32.25          &0.909      &0.019     &5.335     &71.55               &0.398     &22.91     &137.58     \\
        \hline
\rowcolor[HTML]{EFEFEF}       DB-CR & \textbf{33.47}          & \textbf{0.922}      & \textbf{0.016}       & \textbf{4.740}   & \textbf{61.91}   & \textbf{0.316}               & 18.06      & \textbf{15.24} \\
\bottomrule
\end{tabular}}
\label{tb:main}
\end{table*}

\subsection{Comparison with baseline methods}
We selected several well-known methods as references to compare the performance of DB-CR:
\begin{itemize}
\item \textbf{DSen2-CR}~\cite{dsen2-cr}: Dsen2-CR method removes clouds from Sentinel-2 satellite images using a deep residual neural network in which the input is a concatenation of the optical and SAR images.
\item \textbf{GLF-CR}~\cite{glf-cr}: Global-Local Fusion Cloud Removal (GLF-CR) method utilizes advanced SAR-optical fusion to integrate both global and local feature information from SAR and optical images to estimate the cloud-free image.
\item \textbf{UnCRtainTS}~\cite{uncrtaints}: UnCRtainTS focuses on handling uncertainty in cloud removal tasks, using uncertainty-aware and attention-based fusion techniques to predict cloud-free images from partially cloudy observations.
\item \textbf{DiffCR}~\cite{zou2024diffcr}: DiffCR leverages a diffusion model-based approach to gradually refine cloud removal predictions, integrating a network designed with specialized blocks that fuse temporal and conditional information for enhanced performance. 
\end{itemize}

For a fair comparison, we trained all baseline models with input and output sizes set to $256 \times 256$, using a learning rate of $5 \times 10^{-5}$ and training for 50 epochs on the same training set as our proposed method. For DSen2-CR, we use the same network architecture setting provided by the paper. For GLF-CR, we adjusted the input and output sizes to $256 \times 256$ and conducted full-image training and inference, rather than dividing the images into patches of size of $128 \times 128$ as in the original implementation, to avoid any performance differences stemming from patch-based processing. For UnCRtainTS, the original model was designed with a lightweight purpose in mind. To prioritize performance, we scaled up the model parameter size from 0.5M to 2.02M -- its performance did not continue to improve with larger sizes -- and is used as the baseline. For DiffCR, we followed the original implementation and selected the Number of Function Evaluations (NFE) at inference time equal to~3, which provided the best PSNR.

\begin{figure*}
	\centering
	\includegraphics[width=\textwidth]{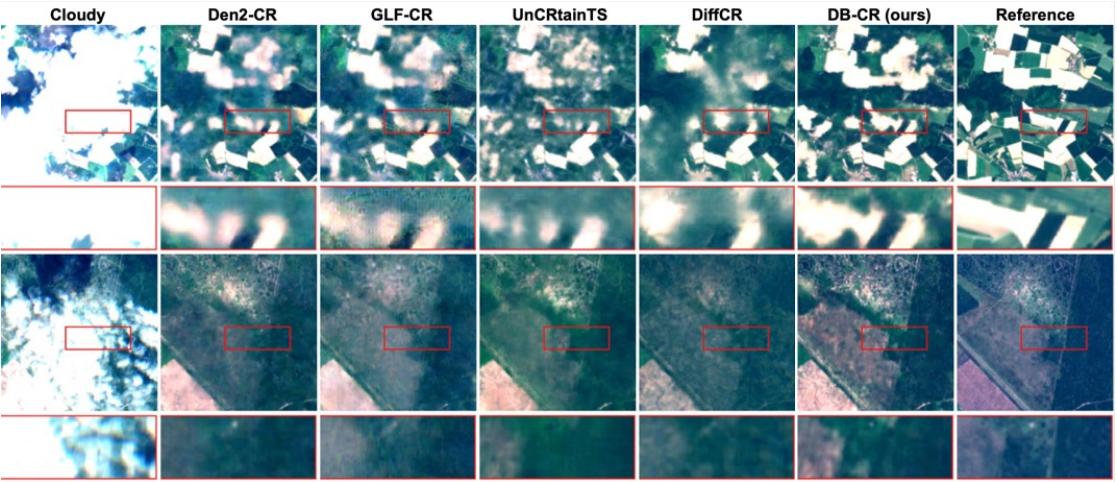}
	\caption{Comparison of DB-CR vs. baseline methods on two agricultural scenes. For each image in the first and third rows, the detail region enclosed in the red rectangle is magnified in the subsequent row. For each scene, the original cloudy image is shown in the leftmost column, and the reference image (actual cloud-free image of the same area as the cloudy image) is in the rightmost column. Compared to the baseline methods, DB-CR achieves finer detail recovery, accurately preserves edges, and provides sharper details with superior perceptual quality.}
	\label{fig:results_main}
\end{figure*}

\begin{figure*}
	\centering
	\includegraphics[width=\textwidth]{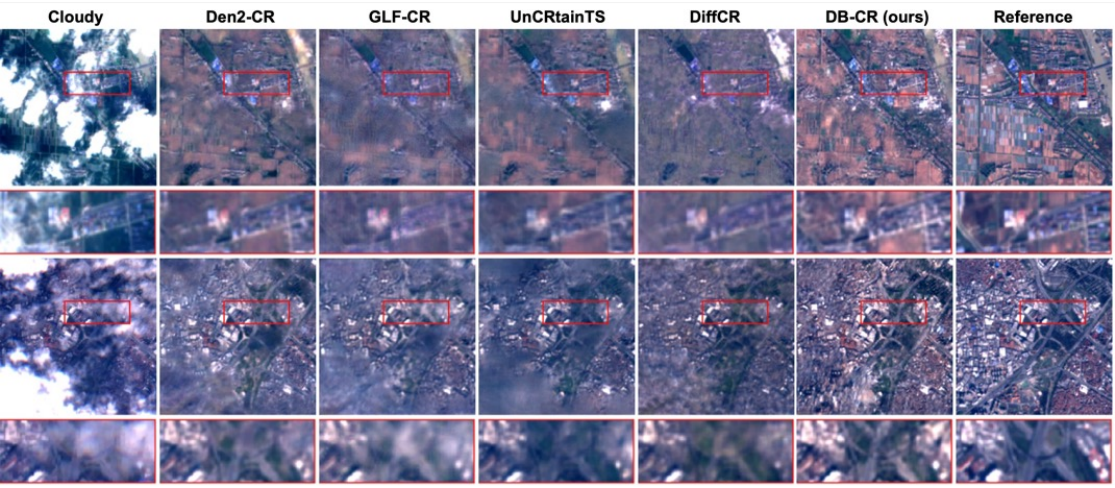}
	\caption{Visual comparison of DB-CR vs. baseline methods on two urban scenes. DB-CR demonstrates superior detail recovery and effectively addresses the oversmoothing issues that plague all of the baseline methods.}
	\label{fig:results_main_city}
\end{figure*}

Table~\ref{tb:main} presents the quantitative results of all evaluated methods. We use NFE=1 for DB-CR. As shown, DB-CR consistently outperforms the others, achieving the highest PSNR and SSIM scores, and the lowest MAE and SAM values. In addition, DB-CR exhibits the best perceptual quality, reflected in the best FID and LPIPS scores. Notably, our method achieves this superior performance with a comparable number of parameters to the baseline methods, while significantly reducing MACs, which translates to faster speeds in each inference step. In Figures~\ref{fig:results_main} and ~\ref{fig:results_main_city}, we show some qualitative results with the RGB channels of the multispectral images visualized. Again, results obtained DB-CR are of signficantly higher quality compared to the baselines.

We further evaluate our method’s performance across varying cloud cover levels. Using the S2Cloudless cloud detector~\cite{s2cloudless}, we generated binary masks to identify cloud pixels and calculated the percentage of cloud cover in each test image. We then computed the median quantitative metrics at 5 different cloud cover levels, ranging from 0–20\% to 80–100\%. As shown in Table~\ref{tb:cloud_cover}, our method consistently outperforms DiffCR across all cloud cover levels, demonstrating robust performance even under heavy cloud coverage.

\begin{table*}[hbt!]
\centering
\caption{Effect of backbone architecture and fusion method on cloud removal performance. We see that the proposed combination of NAFBlocks in the backbone and attention-based SAR-Optical feature fusion results in the best cloud removal performance.}
\renewcommand\arraystretch{1.25}
\setlength{\tabcolsep}{1mm}{
    \begin{tabular}{@{}c@{\hskip 5pt}|@{\hskip 5pt}c@{\hskip 5pt}c@{\hskip 5pt}c@{\hskip 5pt}c@{\hskip 5pt}c@{\hskip 5pt}c@{\hskip 5pt}c@{\hskip 5pt}c@{}}
\toprule
Backbone &Fusion Method & PSNR(dB) $\uparrow$ & SSIM $\uparrow$ & MAE $\downarrow$ & SAM $\downarrow$ & FID $\downarrow$ & LPIPS $\downarrow$\\ \hline
        ResBlock  & Early concatenation &30.37         &0.879      &0.032	     &6.972    &79.50	     &0.482     \\
        NAFBlock  & Early concatenation &32.75         &0.916      &0.018	     &5.190   &70.11	     &0.368  \\
        ResBlock  & Multi-depth cross-modal attention &31.95         &0.894      &0.024	     &6.251    &74.69	     &0.437    \\
        \hline
\rowcolor[HTML]{EFEFEF}       NAFBlock  & Multi-depth cross-modal attention &\textbf{33.47}          &\textbf{0.922}      &\textbf{0.016}       &\textbf{4.740}   &\textbf{61.91}   & \textbf{0.316} \\ \bottomrule
\end{tabular}}
\label{tb:ablation_network}
\end{table*}

\begin{table*}[hbt!]
\centering
\caption{Comparison of DB-CR with baseline method DiffCR (a conditional diffusion model) under different cloud-cover percentages. Our proposed DB-CR method outperforms DiffCR for all cloud cover amounts.}
\renewcommand{\arraystretch}{1.3}
\setlength{\tabcolsep}{1.5mm}{
\begin{tabular}{c c c c c c c c c c}
\toprule
\textbf{Cloud Cover (\%)} & \textbf{Method} & \textbf{PSNR} $\uparrow$ & \textbf{SSIM} $\uparrow$ & \textbf{MAE} $\downarrow$ & \textbf{SAM} $\downarrow$ & \textbf{FID} $\downarrow$ & \textbf{LPIPS} $\downarrow$ \\ 
\midrule
\multirow{2}{*}{0\%-20\%}  & DiffCR & 35.07 & 0.956 & 0.011 & 2.82 & 56.15 & 0.203\\
                           & DB-CR & \textbf{36.04} & \textbf{0.964} & \textbf{0.010} & \textbf{2.71} & \textbf{44.73} & \textbf{0.150} \\ 
\cmidrule{2-8}
\multirow{2}{*}{20\%-40\%} & DiffCR & 32.61 & 0.944 & 0.015 & 3.47 & 68.76 & 0.279\\
                           & DB-CR & \textbf{33.92} & \textbf{0.953} & \textbf{0.013} & \textbf{3.08} &\textbf{62.46} & \textbf{0.237} \\ 
\cmidrule{2-8}
\multirow{2}{*}{40\%-60\%} & DiffCR & 32.48 & 0.926 & 0.015 & 4.02 & 76.88 & 0.324\\
                           & DB-CR  & \textbf{33.55} & \textbf{0.937} & \textbf{0.013} & \textbf{3.52} &\textbf{70.54} & \textbf{0.291}\\ 
\cmidrule{2-8}
\multirow{2}{*}{60\%-80\%} & DiffCR & 31.90 & 0.906 & 0.016 & 4.46 & 95.26 & 0.362\\
                           & DB-CR  & \textbf{32.83} & \textbf{0.918} & \textbf{0.015} & \textbf{4.02} &\textbf{83.61} & \textbf{0.331}  \\ 
\cmidrule{2-8}
\multirow{2}{*}{80\%-100\%} & DiffCR & 30.85 & 0.888 & 0.019 & 5.73 & 123.31 & 0.446\\
                            & DB-CR  & \textbf{31.62} & \textbf{0.903} & \textbf{0.017} & \textbf{5.09} &\textbf{106.98} & \textbf{0.407} \\ 
\bottomrule
\end{tabular}}
\label{tb:cloud_cover}
\end{table*}

\subsection{Ablation study for proposed DB-CR backbone architecture}
This section presents the ablation study results for the DB-CR backbone architecture. As described in Section~\ref{Sec:Method}, we employ a two-branch U-Net architecture as the core of our model. To enhance both the speed and quality of feature extraction and improve SAR-optical image fusion, we introduced two key modifications: time-embedded NAFBlocks and SFBlocks.

As shown in Table~\ref{tb:ablation_network}, the base model, which uses a two-branch U-Net with traditional residual blocks and simple concatenation (``early concatenation'') for SAR-optical feature fusion, serves as the reference. Replacing the residual blocks with NAFBlocks in the backbone results in a significant improvement in reconstruction accuracy, reflected in the increased PSNR and SSIM values and the reduction in error metrics. Similarly, introducing SFBlocks (``multi-depth cross-modal attention'') enhances multimodal fusion, leading to gains in performance. The combination of both NAFBlocks and SFBlocks yields the most substantial improvements, confirming their complementary roles in refining cloud removal and improving the fidelity of the final cloud-free images.

\section{Discussion}
\subsection{Comparison with diffusion model} 
Our proposed method is based on a diffusion bridge, and our results show that this is more effective at solving the cloud removal problem than a standard conditional diffusion model~\cite{zou2024diffcr}. Traditional diffusion models employ a forward process that progressively adds noise to an image and learn a reverse process to denoise it, ultimately generating a clean version of the image. This process can be effective for image synthesis and some restoration tasks, but it often lacks control over the specific transition between distinct input and output states, such as in the cloud-removal task where the input (cloudy) and output (cloud-free) states are well defined.

In contrast, our diffusion bridge explicitly models the transition between the cloudy image and its cloud-free counterpart, creating a targeted path that connects these two states. This allows for a more direct and constrained approach to the cloud-removal problem, ensuring that each step in the denoising process moves the image closer to a plausible cloud-free version. This characteristic provides a significant advantage over standard diffusion models, which may diverge from the desired solution due to the randomness in the reverse (denoising) process. By explicitly guiding the transition from cloudy to cloud-free states, the diffusion bridge achieves a more consistent and stable reconstruction, with reduced risk of oversmoothing or artifacts.

Moreover, the diffusion bridge formulation allows for the use of optimal transport to guide the intermediate states in a way that directly reflects the transition from cloudy to cloud-free. This more focused approach results in improved consistency and perceptual quality of the final output, reducing the risk of oversmoothing or introducing artifacts commonly seen in standard diffusion models. 

Ultimately, the direct nature of the diffusion bridge makes it better suited for cloud removal, where a specific target state needs to be recovered accurately and efficiently.

\subsection{Advantage of DB-based training for one-step inference}
To further understand the benefits of our approach, we performed a comparison with a one-step end-to-end learning method that employs the same network architecture and SAR fusion as DB-CR. This ablation study method uses a single-step mapping ($T=1$) from the cloudy image to its reference during both training and inference. In contrast, our proposed method (DB-CR) uses multiple gradual degradation levels (e.g., 1000 levels) for training but can perform inference in a single step by setting NFE = 1. Importantly, when setting NFE to 1, DB-CR essentially reduces to a single-step direct estimation from the cloudy image to its reference, achieving the same inference speed as its end-to-end learned counterpart. As demonstrated in Table~\ref{tb:ablation_db}, this single-timestep method significantly underperformed compared to our proposed strategy, which uses a more dynamic timestep size during training. We believe this discrepancy is due to the dynamic timestep acting as a form of data augmentation, making the model more robust to varying cloud conditions. By randomly sampling different timesteps during training, the model is effectively exposed to a range of intermediate states of degradation, allowing it to learn how to adapt to different cloud-cover levels. This variability enables the model to develop a more generalized understanding of the problem, akin to data augmentation, which is crucial for handling the diverse nature of cloud coverage. This adaptability ensures that the model can apply the appropriate level of reconstruction precision across different regions of an image, ultimately resulting in superior cloud-free outputs.

\begin{table}[ht]
\centering

\caption{Ablation study of DB-CR for diffusion bridge training. Here, ``w/ DB'' refers to our full DB-CR method. In constrast, ``w/o DB'' refers to end-to-end training of the backbone without time embedding. All other hyperparameters are the same as for DB-CR}
    \begin{tabular}{@{}c@{\hskip 5pt}|@{\hskip 5pt}c@{\hskip 5pt}c@{\hskip 5pt}c@{\hskip 5pt}c@{\hskip 5pt}c@{\hskip 5pt}c@{}}
    \toprule[1pt]
    Method  & PSNR $\uparrow$ & SSIM $\uparrow$ & MAE $\downarrow$ & SAM $\downarrow$   & FID $\downarrow$ & LPIPS $\downarrow$ \\ \midrule
    w/o DB   &33.03          &0.916      &0.017     &5.074 &{66.08}   & {0.335}\\
    w/ DB   &\textbf{33.47}          &\textbf{0.922}      &\textbf{0.016}      &\textbf{4.740} &\textbf{61.91}   & \textbf{0.316}\\ 
    \bottomrule[1.2pt]
    \end{tabular}
    \label{tb:ablation_db}

\end{table}

 \begin{figure*}[ht]
	\centering
	\includegraphics[width=\textwidth]{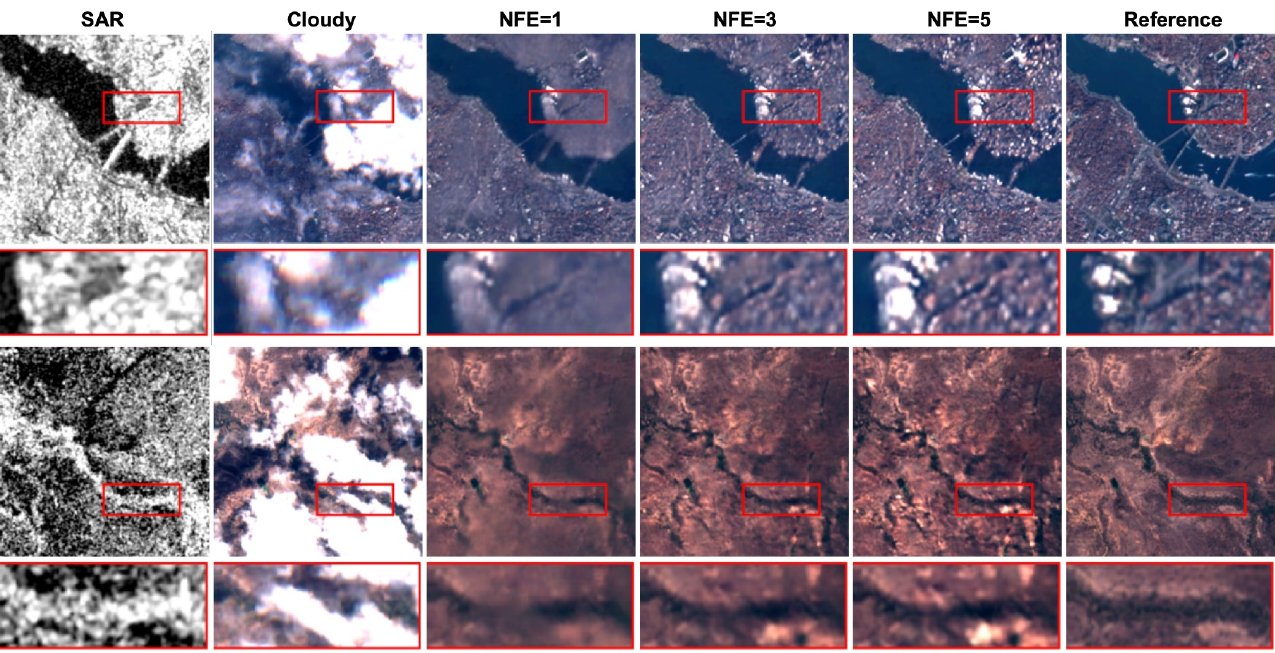}
	\caption{Impact of NFE (number of function evaluations for inference) on DB-CR outputs. Higher NFE enhances perceptual quality by preserving finer details, while lower NFE provides smoother prediction, balancing visual clarity and data consistency.}
	\label{fig:nfe}
\end{figure*}

\subsection{NFE-controlled distortion-perception trade-off} 
Previous work~\cite{blau2018perception} has established that there is an inherent trade-off between image perceptual quality and distortion in reconstruction tasks, making it impossible to minimize both simultaneously. As described in Equation~\eqref{eq:inference_ode}, the selection of timesteps during the inference phase introduces a degree of freedom that allows us to control the NFE (number of function evaluations for inference). According to InDI~\cite{indi2023}, different NFE values allow for a controllable distortion-perception trade-off, which is particularly significant for cloud-removal (CR) tasks. Motivated by this, we explore how the number of evaluation steps affects the multi-step restoration process in the diffusion bridge framework.

For cloud removal, reducing the NFE yields estimates that are closer to the posterior mean, effectively minimizing distortion but potentially causing over-smoothing, which can result in a loss of important details. In contrast, increasing the NFEs enhances perceptual quality, making the reconstructed images appear more visually natural, but also introduces a risk of generating unrealistic details.  As shown in Table~\ref{tb:ablation_nfe}, the choice of NFE has a direct impact on the trade-off between distortion metrics (PSNR, SSIM, MAE, SAM) and perceptual metrics (FID, LPIPS). Higher NFEs improve perceptual quality at the expense of distortion accuracy, while lower NFEs enhance distortion performance but reduce perceptual fidelity. According to Table~\ref{tb:ablation_nfe}, setting NFE $=3$ provides a favorable balance between PSNR and LPIPS scores, preserving cloud-free accuracy while maintaining good perceptual quality. Furthermore, observing results in Table~\ref{tb:main}, running DB-CR with both NFE $=1$ and NFE $=3$ configurations outperform all baseline methods across distortion and perceptual metrics.

For applications requiring higher visual fidelity, increasing NFE can further improve perceptual quality, albeit at the cost of increased distortion. For instance, as illustrated in the top row of Figure~\ref{fig:nfe}, higher NFE results in more distinct gaps appearing in the terrain rather than smooth connections. Similarly, in the bottom row, higher NFE makes it increasingly clear that the area beneath the cloud contains buildings, as opposed to the blurry artifacts seen with NFE $=1$.

\subsection{Influence of stochastic perturbation for CR} We investigate the impact of introducing stochastic perturbation into the problem formulation. In general image restoration tasks, the effect of stochastic perturbation tends to be task-dependent. For instance, in JPEG restoration, stochastic perturbation has been shown to enhance performance, whereas in deblurring, it often leads to degradation in quality~\cite{indi2023, liu2023i2sb}. Therefore, it is valuable to evaluate the influence of stochastic perturbation on the cloud removal task.

In our formulation, stochastic perturbation can effectively convert the ODE-based formulation of DB-CR into an SDE-based approach. Specifically, we modify the forward process of the diffusion bridge as follows:
\[
\xbm_t = (1 - \alpha_t)\xbm_0 + \alpha_t \ybm + \beta_t\epsilon, \quad \alpha_t \in [0, 1], \beta_t \in [0, 1] , 
\]
where $\beta_t\epsilon_t$ is the standard Brownian motion having zero mean and covariance
$\beta_t\Ibf$ at time $t$. In this SDE-based approach, the DB network is also trained to perform MMAE estimation ($\E[\xbm_{0}|\xbm_{t}]$) of the target cloud-free image using the same training strategy as the ODE-case in Equation~\eqref{eq:train_ode_loss}. For inference, however the SDE case will be different from the ODE case shown in Equation~\eqref{eq:inference_ode}. Instead, inference in the SDE case uses:
\begin{equation}
\label{eq:inference_sde}
\begin{matrix}
        \E[\xbm_{t-1}|\xbm_{t}] = \left(1 - \frac{\alpha_{t-1}}{\alpha_t}\right)\E[\xbm_{0}|\xbm_{t}] + \left(\frac{\alpha_{t-1}}{\alpha_t}\right) \xbm_t \\ \hspace{100pt} +\left(\beta_t\frac{{\alpha}_{t-1}}
     {\alpha_t}-\beta_{t-1} \right)\epsilon. 
\end{matrix}
\end{equation}

\begin{table}[ht]
\caption{Effect of Number of Function Evaluations (NFE) through the DB-CR backbone on cloud removal performance. Distortion metrics are best with NFE$=1$, while perceptual metrics improve with higher NFE. This confirms the well-known perception-distortion trade-off in image restoration.}
\centering
    \begin{tabular}{c|cccccc}
    \toprule[1pt]
    NFE  & PSNR $\uparrow$ & SSIM $\uparrow$ & MAE $\downarrow$ & SAM $\downarrow$ & FID $\downarrow$ & LPIPS $\downarrow$    \\ \midrule
    1   &\textbf{33.47}          &\textbf{0.922}      &\textbf{0.016}       &\textbf{4.740} &61.91 &0.316\\
    3   &33.10          &0.915      &0.017       &5.031 &\textbf{58.03} &0.273\\
    5   &32.51          &0.905      &0.018       &5.288 &58.40 &\textbf{0.269}\\
    
    \bottomrule[1.2pt]
    \end{tabular}
    \label{tb:ablation_nfe}

\end{table}
\begin{table}[ht]
\centering
\caption{Performance comparison of DB-CR with SDE and ODE formulation.}
\begin{tabular}{@{}c@{\hskip 5pt}|@{\hskip 5pt}c@{\hskip 5pt}c@{\hskip 5pt}c@{\hskip 5pt}c@{\hskip 5pt}c@{\hskip 5pt}c@{}}    \toprule[1pt]
    Method  & PSNR $\uparrow$ & SSIM $\uparrow$ & MAE $\downarrow$ & SAM $\downarrow$ & FID $\downarrow$ & LPIPS $\downarrow$  \\ \midrule
    DB-CR (SDE)   &32.95 &0.914 &0.018     &5.133 &{68.51}   & {0.369}\\
    DB-CR (ODE)   &\textbf{33.47}          &\textbf{0.922}      &\textbf{0.016}       &\textbf{4.740} &\textbf{61.91}   & \textbf{0.316}\\ 
    \bottomrule[1.2pt]
    \end{tabular}
    \label{tb:ablation_sde}

\end{table}

As shown in Table~\ref{tb:ablation_sde}, our experiments indicate that the SDE-based approach is not well-suited for the cloud removal problem, leading to reduced performance in both distortion and perception metrics. The added randomness, while theoretically capable of preventing over-smoothing and encouraging diverse solutions, often results in inconsistent output quality. In practice, the cloud removal task benefits more from a deterministic approach where the reconstruction process is guided precisely to recover missing details without introducing additional uncertainty. The iterative refinement in the ODE-based method ensures a more controlled transition from the cloudy to the clean state, leading to better preservation of fine details and overall perceptual quality.

\section{Conclusion}
In this paper, we introduced DB-CR, a novel diffusion bridge-based method designed for cloud removal. DB-CR leverages fixed start and end conditions to guide a precise, step-by-step reduction of cloud cover, closely aligning with the desired cloud-free state. By applying optimal transport principles, DB-CR achieves an optimal balance of fidelity and efficiency at each intermediate step, preserving structural integrity and perceptual quality throughout the process. Our diffusion bridge-based network backbone, coupled with the fusion of features from both the SAR and optical modalities, establishes DB-CR as the leading approach for both accuracy and computational efficiency. Experimental results validate DB-CR’s state-of-the-art performance in terms of both distortion metrics and perceptual quality.

{\small
	\bibliographystyle{IEEEtran}
	\bibliography{strings}
}

\end{document}